\documentclass[10pt,twocolumn,letterpaper]{article}

\usepackage{cvpr}
\usepackage{times}
\usepackage{graphicx}
\usepackage{amsmath}
\usepackage{amssymb}
\usepackage{amsfonts}
\usepackage{multirow}
\usepackage{here}
\usepackage{bm}
\usepackage{enumerate}
\usepackage{tabularx}
\usepackage{xcolor}
\usepackage{url}
\usepackage{color}
\usepackage{longtable}
\usepackage[subrefformat=parens]{subcaption}
\usepackage{algorithm}
\usepackage{algorithmic}
\usepackage{threeparttable}
\usepackage{cuted}
\usepackage{capt-of}
\usepackage{nidanfloat}

\def\rnum#1{\expandafter{\romannumeral #1}}
\def\Rnum#1{\uppercase\expandafter{\romannumeral #1}}
\newcommand{\argmax}{\mathop{\rm arg~max}\limits}
\newcommand{\argmin}{\mathop{\rm arg~min}\limits}
\DeclareMathOperator{\atantwo}{atan2}

\usepackage[breaklinks=true,letterpaper=true,colorlinks,bookmarks=false]{hyperref}
\cvprfinalcopy  
\def\cvprPaperID{7104} 

\begin{document}
\title{
\vspace{-0.8cm}
Synergetic Reconstruction from 2D Pose and 3D Motion \\
for Wide-Space Multi-Person Video Motion Capture in the Wild
}
\let\SUP\textsuperscript
\author{
{\large Takuya Ohashi\SUP{1,2} \hspace{1cm} Yosuke Ikegami\SUP{2} \hspace{1cm} Yoshihiko Nakamura\SUP{2}}\\
{\SUP{1}NTT DOCOMO \hspace{1cm} \SUP{2}The University of Tokyo}\\
{\tt\small takuya.ohashi.ht@nttdocomo.com} \hspace{0.1cm} {\tt\small ikegami@ynl.t.u-tokyo.ac.jp} \hspace{0.1cm} {\tt\small nakamura@race.t.u-tokyo.ac.jp}
}

\makeatletter
\let\@oldmaketitle\@maketitle
\renewcommand{\@maketitle}{\@oldmaketitle
  \vspace{-0.8cm}
  \includegraphics[width=\linewidth]{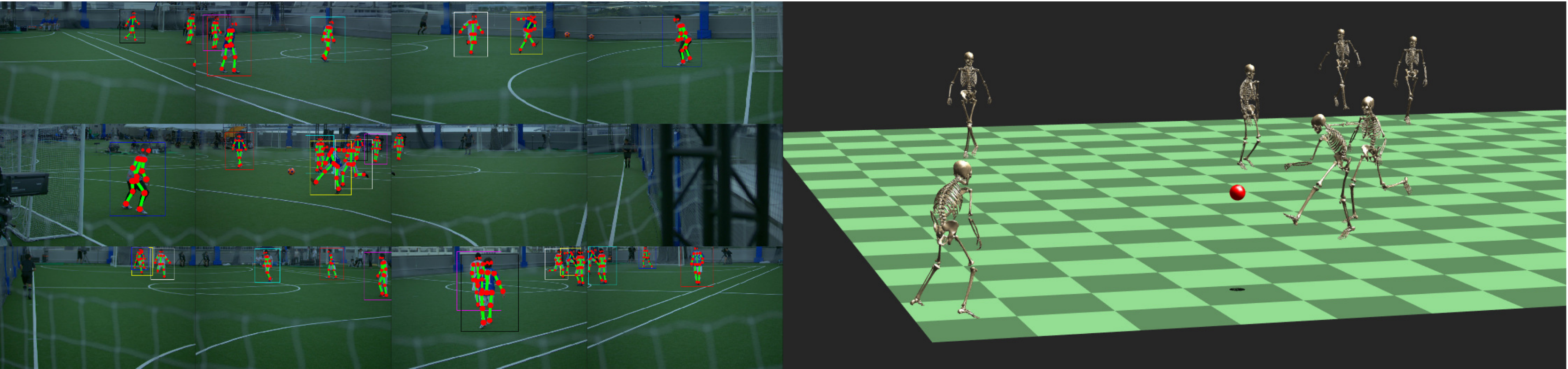}
  \captionof{figure}{All futsal players' motions were captured using 12 video cameras surrounding the court. (left) Input images and reprojected joint position. (right) Bone CG drawing based on the calculated joint angles.
  \label{fig:feature}}
  \vspace{0.2cm}}
\makeatother

\maketitle

\begin{abstract}
Although many studies have investigated markerless motion capture, the technology has not been applied to real sports or concerts.
In this paper, we propose a markerless motion capture method with spatiotemporal accuracy and smoothness from multiple cameras in wide-space and multi-person environments.
The proposed method predicts each person's 3D pose and determines the bounding box of multi-camera images small enough.
This prediction and spatiotemporal filtering based on human skeletal model enables 3D reconstruction of the person and demonstrates high-accuracy.
The accurate 3D reconstruction is then used to predict the bounding box of each camera image in the next frame.
This is feedback from the 3D motion to 2D pose, and provides a synergetic effect on the overall performance of video motion capture.
We evaluated the proposed method using various datasets and a real sports field.
The experimental results demonstrate that the mean per joint position error (MPJPE) is 31.5 mm and the percentage of correct parts (PCP) is 99.5\% for five people dynamically moving while satisfying the range of motion (RoM). Video demonstration, datasets, and additional materials are posted on our project page\footnote{\scriptsize{\label{note1}\url{http://www.ynl.t.u-tokyo.ac.jp/research/vmocap-syn}}}.
\end{abstract}

%
%

\section{Introduction}

Human motion data are widely used in various fields, e.g., sports training, CG production, rehabilitation, medical diagnosis, behavioral understanding, and even humanoid robot operation \cite{sdims2,Murai1,takano:2019}.
Various motion capture methods have been developed to obtain such data, e.g., optical motion capture, where reflective markers are attached to characteristic parts of the body, and these 3D positions are then measured \cite{MotionA, Vicon}.
Inertial motion capture uses IMU sensors attached to body parts, and then, the positions are calculated using sensor speed \cite{xsense, neuron}.
Markerless motion capture uses a depth camera or single/multiple RGB video cameras \cite{Shotton1, Tong1, radical, captury}.
However, although various methods for using motion data exist, this technology is only used in limited locations.
Few examples of motion capture being used in locations with practical value (e.g., sports matches, concerts, and public roadways) have been reported.

This encourages the question ``why are motion data not captured in the real world?'' Motion capture under real-world conditions is challenging because human motion is continuous; thus, the motion data must also be continuous, i.e., parts or all of the body must not
be lost at any time. However, the real world has three specific factors that make
motion capture difficult.
The first difficulty is the existence of multiple subjects, which causes occlusion and requires individual identification and tracking.
The second difficulty is related to the large measurement field.
A wider measurement field incurs greater calibration
error; however, precise calibration is required in motion capture. In addition, the measurement field
can sometimes be open, i.e., people can enter and exit the field.
The third difficulty is derived from real-world environments, which are not ideal and restrict measurement conditions. For competitive
sporting events or concerts, measurement constraints must be avoided,
e.g., markers, IMU sensors, or specific shirts/pants. Furthermore, other
constraints exist, e.g., taking measurements in a severe lighting
conditions or being unable to set the sensor at the desired position.
Due to these various difficulties, even with the latest technology, motion capture under real-world conditions has not been fully developed.

In this paper, we discuss the multi-person video motion capture, which means image-based 3D human motion reconstruction with spatiotemporal accuracy and smoothness even in a challenging multi-person environment, by extending the single-person video motion capture method \cite{Ohashi:2018}.
In the proposed method, multiple synchronized calibrated cameras are used to record video images of human subjects from different directions.
A human skeletal model is also used to reconstruct 3D motion by spatiotemporal filtering of joint movements. The key concept of the proposed method is predicting each person's 3D pose and determining the bounding box small enough. Using this bounding box, the keypoint positions of each subject in each image are estimated using a top-down pose estimation approach \cite{Xiao:2018, Sun:2019}.
The estimated positions are received as part confidence maps (PCM) which express the probability of the keypoint existence at each pixel location as continuous values in the range [0, 1].
Probable keypoint positions can be calculated using the PCM of multi-camera images and a predicted past 3D motion.
Then, the skeletal model's current 3D pose is reconstructed by minimizing the error between the probable keypoint positions and the skeletal model's corresponding joint positions.
The reconstructed 3D motion is then used to predict the bounding box of each camera image in the next frame.
This feedback from 3D motion to 2D pose provides a synergetic effect on the overall video motion capture performance.

The proposed method was quantitatively evaluated using various datasets \cite{shelf}\textsuperscript{\ref{note1}}. We also applied the proposed method to actual futsal matches to evaluate it in real-world environments.
Additionally, the proposed method uses inverse kinematics (IK) for optimization; thus, it is possible to calculate not only the position but joint angle considering the range of motion (RoM). As a qualitative evaluation, bone CG was generated using the joint angle as shown in Fig. \ref{fig:feature}.

\section{Related work}
\subsection{Single-view pose estimation}
Human 2D pose estimation from a single image is a task of detecting human keypoint positions, e.g., knees and shoulders in an image.
Typically, two approaches are used: the top-down approach, which first detects the positions of multiple
people in an image as a bounding box and then estimates the keypoint positions of a single person in the cropped image \cite{2017MaskR, Chen:2018, Xiao:2018, Sun:2019}, and the bottom-up approach, which first estimates the 2D keypoint positions of all people in the entire image and then
associates the positions for each person \cite{Wei:2016, Cao:2017, Kreiss:2019, higherHRNet}.
In general, top-down approaches are more accurate, and bottom-up approaches are faster.
However, a top-down approach is heavily dependent on human detection results for accuracy; therefore, estimation is likely to fail in environments with severe occlusion.

In recent years, several studies have estimated human 3D poses only from a single image by extending detected 2D keypoint positions to 3D spaces \cite{Moreno:2017, Martinez:2017, Akhter2015}, directly estimating 3D poses \cite{VNect:2017,Bogo:2016,Kanazawa1,Zhou_2017_ICCV,singleshot}, and estimating not only poses but detailed body shapes \cite{Xiang:2019, livecap:2019}.
However, 3D pose estimation from a single image is a fundamentally ill-posed problem because various assumptions must be made.
Therefore, the estimation accuracy obtained in a complex environment, e.g., a multi-person environment, is inferior to methods that use multiple cameras.

\subsection{Multi-view 3D pose estimation}

Previous studies have investigated 3D pose estimation using multiple cameras. Most early research efforts extracted a person region from an image, considered the region of the human body in 3D space, and continuously tracked the region over time \cite{Aguiar:2008, Stoll:2011}. This tracking-based approach can independently estimate motion of the subject's pose and has achieved remarkable results.
However, for preparation, it is necessary to create a detailed human model (including clothing). Thus, this approach may fail depending on the light conditions, backgrounds, and clothing of the subject.

In recent years, as 2D pose estimation methods have achieved remarkable results, approaches that combine 2D pose estimation and multi-view geometry have been assessed, e.g., reconstructing estimated keypoints in 3D \cite{Joo:2018, dong2019, Bridgeman} and comparing 3D keypoint probability and 3D pictorial structure \cite{shelf, Belagiannis2, Ershadi2018}. However, most of these methods do not reconstruct 3D keypoint positions when the 2D keypoint is undetected or falsely detected. As a result, continuity, which is essential for motion capture, is lost.
One recent study \cite{4dAssociation} combined per-view parsing, cross-view matching, and temporal tracking and achieved fast, high-performance multi-person motion capture.
However, this study uses a bottom-up pose estimator, so the recognition performance depends largely on the pixel resolution of the person in the image. Therefore, it is difficult to use it in a wide field like a soccer court.

A previous study \cite{Ohashi:2018} proposed a method that uses a bottom-up approach \cite{Wei:2016, Cao:2017} from multiple cameras to estimate 3D keypoint positions, and applies filtering based on the human skeletal model and continuity of joint movements. This method demonstrates high-accuracy and smooth motion capture using a few cameras.
However, this method presented three difficulties.
First, this method specifically examines a single person; thus, in the presence of multiple people, the probability of 3D keypoint positions cannot be computed.
Second, the measurement area is narrow because the area is primarily limited to an overlapping area of four cameras' respective fields of view.
Third, although IK computation is used for filtering, the RoM is not considered. As a result, strange poses may be reconstructed.
We resolve these difficulties and propose a method for high-accuracy and smooth motion capture while satisfying the RoM under multi-person conditions in wide area environments.

\section{Synergetic reconstruction}
The proposed 3D motion reconstruction is performed using $n_c$ synchronized calibrated cameras placed around $n_p$ subjects. During measurements, to avoid difficulties related to a subject cannot be viewed by one camera, multiple cameras with different fields of view are set at a single location.
We designate this location a viewpoint. Here, $n_v$ is the number of viewpoints, $\mathbb{C}_v$ is the set of cameras placed at viewpoint $v$, and $n_{\mathbb{C}_v}$ is the number of cameras at $v$.
\vspace{-0.1cm}
\begin{equation}
n_c = \sum_{v}^{n_{v}} n_{\mathbb{C}_v}
\end{equation}
\vspace{-0.2cm}

\begin{figure*}[!ht]
\centering
\includegraphics[width=\linewidth]{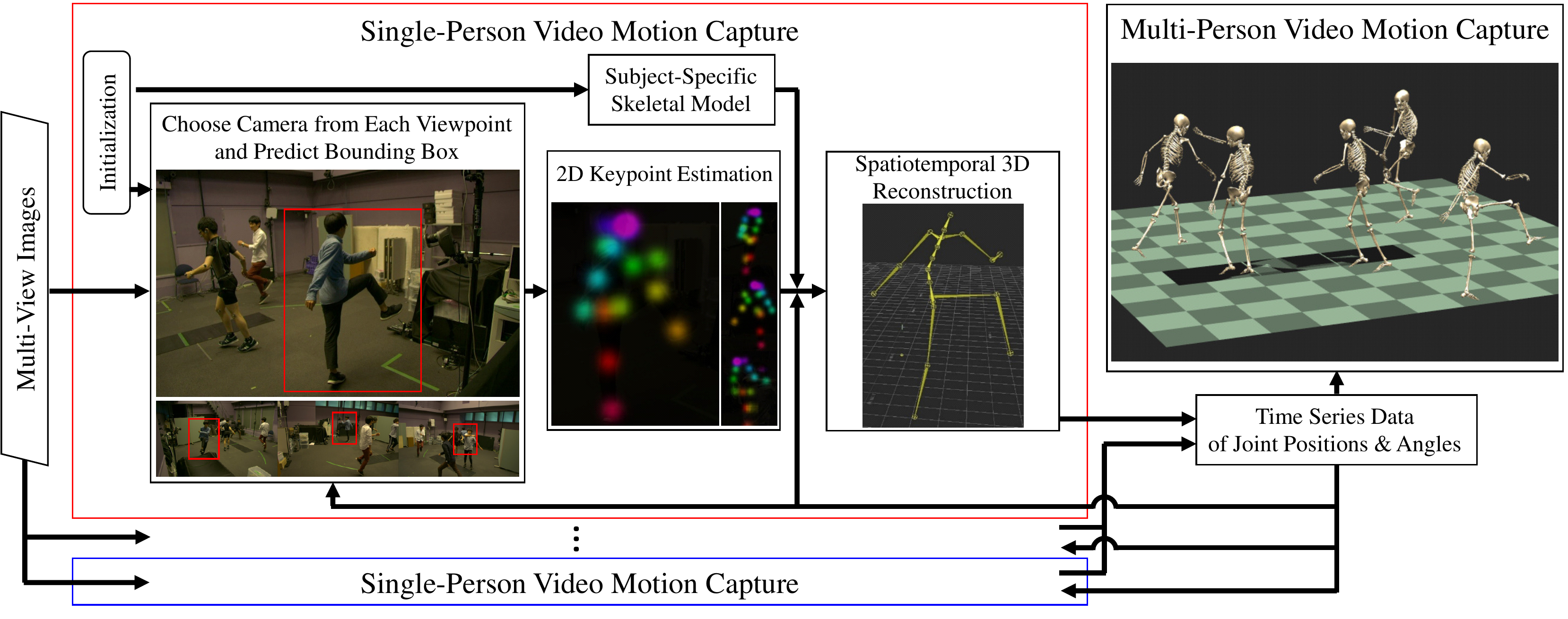}
\caption{Flowchart of proposed multi-person video motion capture method.
\vspace{-0.3cm}
}
\label{fig:flowchart2}
\end{figure*}

\begin{figure}[!ht]
\centering
\includegraphics[width=0.8\linewidth]{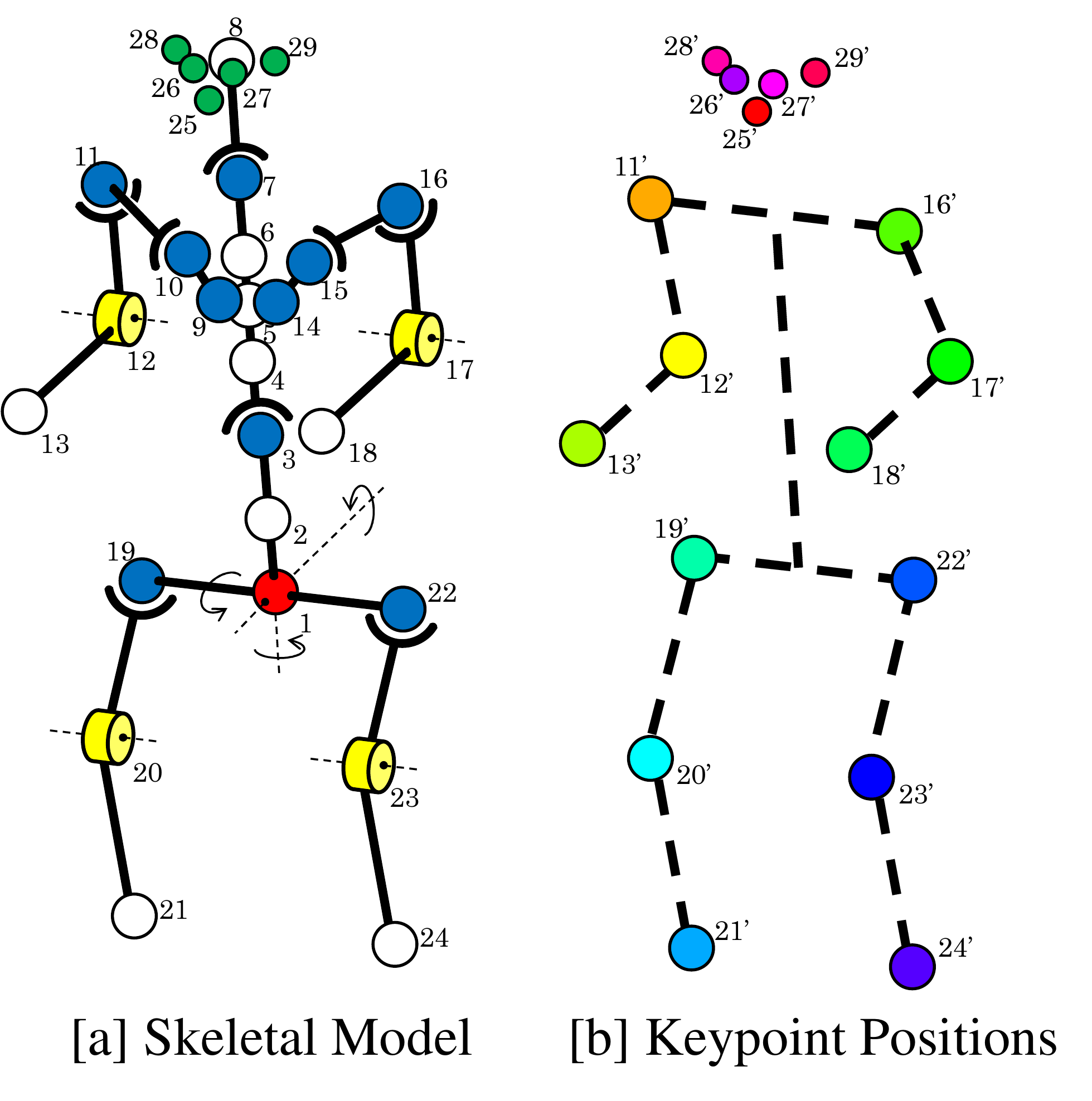}
\caption{Correspondence of human skeletal model and keypoint positions (in [a], red, blue, and yellow represent 6DoF, 3DoF, and 1DoF, respectively).
\vspace{-0.3cm}
}
\label{fig:pcm_figure}
\end{figure}

A flowchart of the proposed method is shown in Fig. \ref{fig:flowchart2}. Each subject's keypoint positions are estimated using a top-down pose estimator, HRNet \cite{Xiao:2018, Sun:2019}. The data are received as PCM. Note that we employ the PCM rather than the pixel location of the keypoint position. With the PCM, we perform spatiotemporal optimization of the human skeletal model and reconstruct the 3D motion. The skeletal model represents a virtual open tree-structure kinematic chain with 40 degrees of freedom (DoF), as shown in Fig. \ref{fig:pcm_figure}[a]. Then, the 3D pose in the next time frame is accurately calculated. The pose is then passed to HRNet as bounding box information. As a result, multi-person video motion capture is realized by applying this process in parallel for each subject and continually repeating the process for each time frame.

Although camera calibration and system initialization (calculating skeletal model's link length and initial joint positions) are important for the proposed method, they are not the primary topics. Therefore, we present details in the \textit{Appendix} and use $\mu _i$ to represent the perspective projection transformation to camera $i$.

\subsection{Determining bounding box from 3D motion}
\begin{figure}[!t]
\centering
\includegraphics[width=\linewidth]{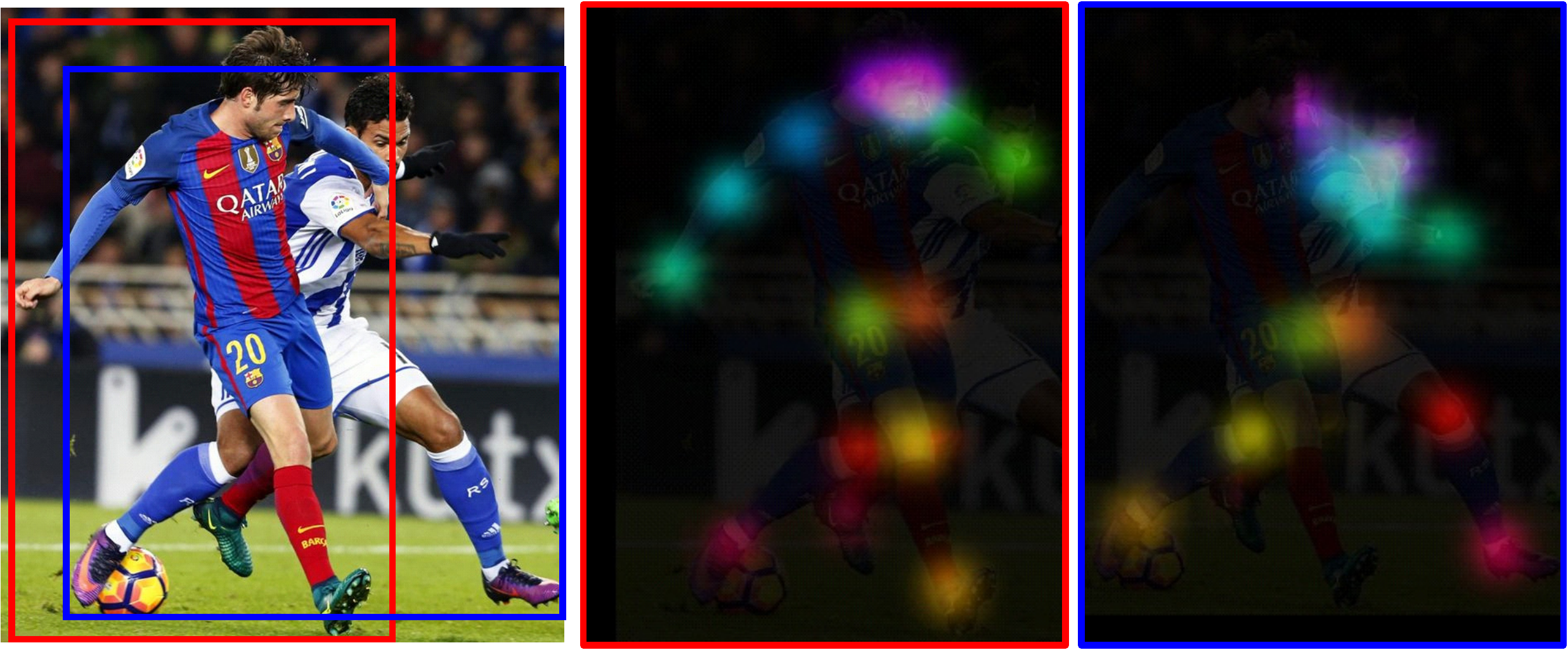}
\caption{2D keypoint estimation using HRNet \cite{Xiao:2018, Sun:2019}. The target person's specific PCM can be estimated by specifying a bounding box for the target person. The input image is from the OCHuman Dataset \cite{pose2seg2019}.
\vspace{-0.3cm}
}
\label{fig:2d_pose}
\end{figure}

In recent years, top-down pose estimation approaches have achieved remarkable results.
If a suitable bounding box is specified, the estimator can robustly and accurately compute only the intended person's PCM, even in severe occlusion environments, as shown in Fig. \ref{fig:2d_pose}.
However, pose estimation in multi-person environments remains challenging.
One factor is that a suitable person region cannot be segmented (e.g., a wrist or ankle is cut).

The proposed method realizes high-accuracy motion capture, and if the frame rate is moderately high, the subject's current 3D pose can be predicted from the calculated past 3D motion.
In addition, the bounding box position can be calculated using perspective projection transformation.
Here, the calculation cost is low. Therefore, we employ the state-of-the-art top-down pose estimation approach: HRNet. The human region is determined from past 3D motion, and the bounding box is simply calculated as follows.
\begin{equation}
\scalebox{0.85}{$
\begin{split}
    _{l}^{t+1} \bf B \it _i =
    \begin{bmatrix}
    \bigl\{\max (\begin{bmatrix} \mu _i ({_{l}^{t+1} \bf P \it _{pred}}) \end{bmatrix}_x) + \min (\begin{bmatrix} \mu _i ({_{l}^{t+1} \bf P \it _{pred}}) \end{bmatrix}_x)\bigr\}/2 \\
    \bigl\{\max (\begin{bmatrix} \mu _i ({_{l}^{t+1} \bf P \it _{pred}}) \end{bmatrix}_y) + \min (\begin{bmatrix} \mu _i ({_{l}^{t+1} \bf P \it _{pred}}) \end{bmatrix}_y)\bigr\}/2 \\
    m\bigl\{\max (\begin{bmatrix} \mu _i ({_{l}^{t+1} \bf P \it _{pred}}) \end{bmatrix}_x) - \min (\begin{bmatrix} \mu _i ({_{l}^{t+1} \bf P \it _{pred}}) \end{bmatrix}_x)\bigr\} \\
    m\bigl\{\max (\begin{bmatrix} \mu _i ({_{l}^{t+1} \bf P \it _{pred}}) \end{bmatrix}_y) - \min (\begin{bmatrix} \mu _i ({_{l}^{t+1} \bf P \it _{pred}}) \end{bmatrix}_y)\bigr\} \\
    \end{bmatrix}
\label{eq:bounding_box1}
\end{split}
$}
\end{equation}
\vspace{-0.2cm}
\begin{equation}
_{l}^{t+1} \bf P \it _{pred} = \rm \frac{3}{2}  \it \hspace{1mm} {_{l}^{t}\bf P \it} - \hspace{1mm} {_{l}^{t-\rm 1 \it}\bf P \it} + \rm \frac{1}{2} \it \hspace{1mm} {_{l}^{t-\rm 2 \it}\bf P \it}
\label{eq:bounding_box2}
\end{equation}
\noindent
Here, $_{l}^{t+1} \bf B \it _i$ represents the predicted center position and size of the bounding box of person $l$ at time $t+1$ for camera $i$, $_{l}^t \bf P$ represents the 3D positions of all joints, and $m$ is a constant positive value whole body becomes just visible.
All joints mean $n_j = 29$ joints, as shown in Fig. \ref{fig:pcm_figure}[a].
Note that assuming uniformly accelerated motion, the future 3D pose is calculated as $^{t+1} \bf P \it = \rm 2 \it \hspace{1mm} {^{t}\bf P \it} - \rm 2 \it \hspace{1mm} {^{t- \rm 1 \it }\bf P \it} + {^{t- \rm 2 \it}\bf P \it}$.
However, we use $^{t+1} \bf P \it _{pred} = { ({^{t+ \rm 1 \it}\bf P \it}+{^{t}\bf P \it}) }/ \rm 2$ as the predicted 3D pose.

For the proposed method, we use a pretrained HRNet model trained on the COCO dataset \cite{coco_dataset}.
The input image is resized and trimmed to $W' \times H' \times 3$ according to the bounding box.
The size of the cropped image is fixed ($W'=288$, $H'=384$), and the PCM is computed from the cropped image.
Here, the number of keypoints is $n_k = 17$, comprising 12 joints (shoulders, elbows, wrists, hips, knees, and ankles) and five feature points (eyes, ears, and nose), as shown in Fig. \ref{fig:pcm_figure}[b].

In addition, HRNet was trained under the assumption that the body is not significantly tilted; thus, estimation may fail when the body is significantly tilted relative to the image's vertical direction, e.g., during a handstand or cartwheel.
With the proposed method, by rotating the bounding box, we can correctly estimate the PCM. The rotation angle is derived from the inclination of the predicted vector connecting the torso and the neck as follows.
\begin{equation}
\scalebox{0.9}{$
\begin{split}
    _{l}^{t+1} \bf B' \rm  \it _i = \frac{\pi}{\rm 2} - \atantwo( & \begin{bmatrix} \mu _i (_{l}^{t+1} P _{pred}^{n(1)}) \end{bmatrix}_y - \begin{bmatrix} \mu _i (_{l}^{t+1} P _{pred}^{n(6)}) \end{bmatrix}_y \\
    &, \begin{bmatrix} \mu _i (_{l}^{t+1} P _{pred}^{n(1)}) \end{bmatrix}_x - \begin{bmatrix} \mu _i (_{l}^{t+1} P _{pred}^{n(6)}) \end{bmatrix}_x)
\label{eq:bb_rotation}
\end{split}
$}
\end{equation}
\noindent
Here, $n$ represents the joint position of the human skeletal model.
This number represents the specific position, as shown in Fig. \ref{fig:pcm_figure}[a]. Note that only 11 keypoints (shoulders, elbows, wrists, eyes, ears, and nose) are calculated from the rotated bounding box.

Also, multiple cameras with different fields of view are set at a single viewpoint; thus, the camera with the greatest visibility of the target person should be selected for 2D keypoint estimation at each viewpoint.
In the proposed method, this selection is performed using the predicted joint position.
\begin{equation}
\begin{split}
    i(v,t,l) = \argmin_{i \in \mathbb{C}_v} & \bigl\{ (\begin{bmatrix} \mu _i (_{l}^{t+1} P _{pred}^{n(1)}) \end{bmatrix}_x - \frac{I_x}{\rm 2})^2 \it \\
    &\quad +(\begin{bmatrix} \mu _i (_{l}^{t+1} P _{pred}^{n(1)}) \end{bmatrix}_y - \frac{I_y}{\rm 2})^2 \it \bigr\}
\label{eq:choose1}
\end{split}
\end{equation}
\noindent
Here, $I$ represents the camera's image resolution.

\subsection{Spatiotemporal 3D motion reconstruction}
To obtain the 3D keypoint position, 3D reconstruction of the detected 2D keypoint position by multiple cameras is conceivable; however, this simple method may fail in severe occlusion environments due to false and missing detections.
Nonetheless, even when the keypoint position is erroneously detected, the PCM may indicate the probability of keypoint existence at the correct keypoint position.
For example, as shown in Fig. \ref{fig:2d_pose}, the PCM of the left ankle of the left person shows the probability at both incorrect and correct positions.
In other words, the PCM is a stochastic field that includes both true positive (TP) and false positive (FP) results.
If only TP results are successfully referenced, then robust 3D reconstruction can be realized in severe occlusion environments.

Here, consider lattice space $_l^{t+1}\mathbb{L}^n$ with $_l^{t+1}P_{pred}^n$ as a center, $s$ as the interval, and $_l^{t+1}L_{a,b,c}^n$ as a single point of the lattice space as:
\begin{equation}
\begin{split}
    _l^{t+1}\mathbb{L}^n :=
     \begin{Bmatrix}
     _l^{t+1}P_{pred}^n + s \left.
    \begin{bmatrix}
    a\\
    b\\
    c\\
    \end{bmatrix} \hspace{0.2cm} \right| \hspace{0.2cm}
    &-k \leq a,b,c \leq k
    \end{Bmatrix}
\label{eq:cube}
\end{split}
\end{equation}
\begin{equation}
    _l^{t+1}L_{a,b,c}^n \in {_l^{t+1}\mathbb{L}^n} ,
\label{eq:cube2}
\end{equation}
\noindent
where $k$ represents constant positive integer, and $a,b,c$ represent integers.
Using perspective projection transformation, one can obtain the PCM value of an arbitrary 3D point at camera $i$. Put simply, if $_l^{t+1}P_{pred}^n$ is accurately predicted, the most probable keypoint position is a point of this grid where the sum of the PCM value is maximum.
This calculation is robust against large false estimation error and lighter than considering a huge stochastic field by projecting multiple PCMs into 3D space.

\begin{figure}[!t]
\centering
\includegraphics[width=\linewidth]{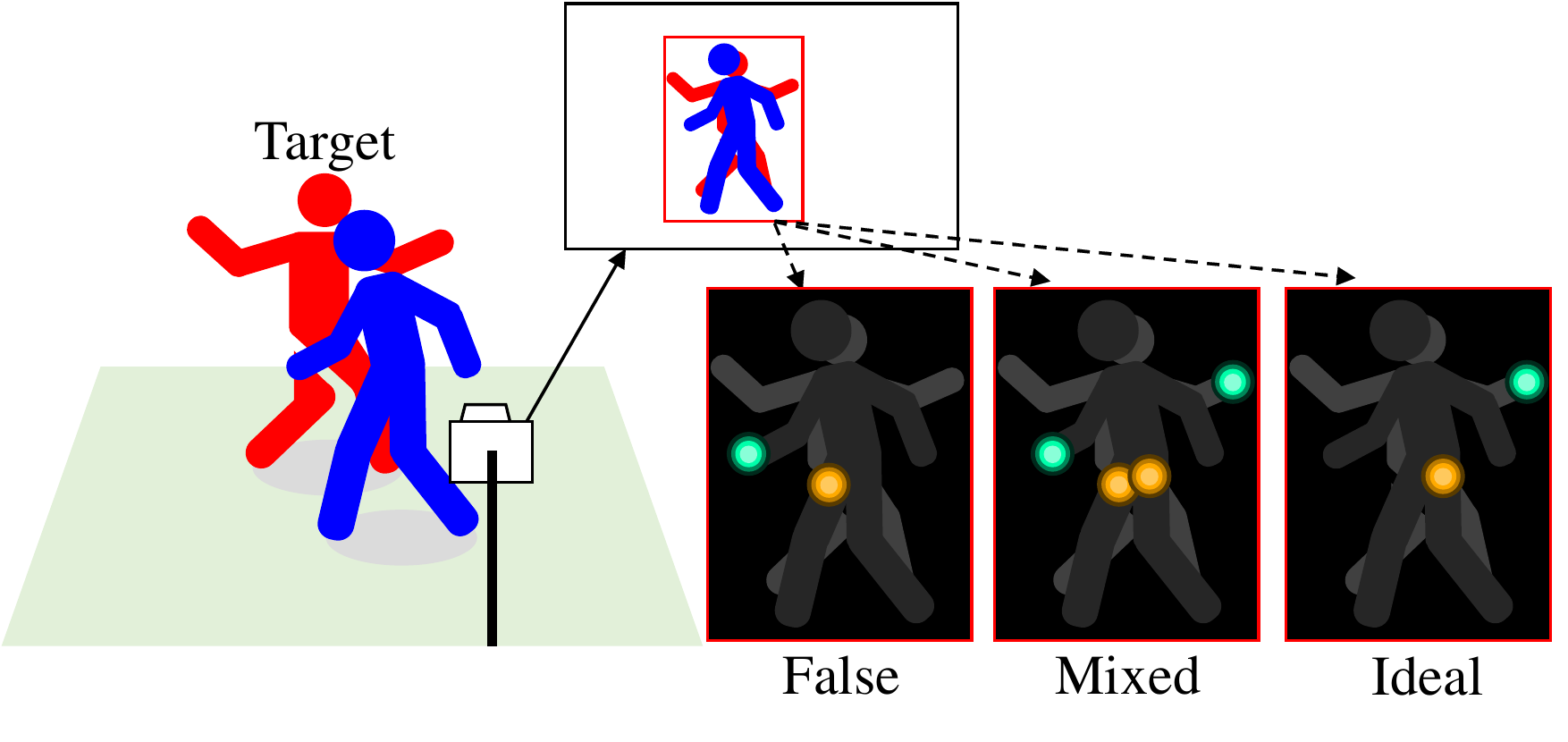}
\caption{PCM computation with truly severe occlusion. Here, the pose estimator estimates the keypoints of the right wrist and right hip of the red person, but the actual results are unknown.
\vspace{-0.3cm}
}
\label{fig:occlusion}
\end{figure}

However, the proposed method targets multi-person environments. The top-down approach attempts to compute the PCM of the intended person in the bounding box; however, this approach suffers some limitations. For example, it may compute unintended PCM if truly severe occlusion occurs as shown in Fig. \ref{fig:occlusion}.
However, the PCM computation in such an occlusion environment is difficult to quantitatively treat. Even under similar environments, various estimation results can be obtained, e.g., false, mixed, and ideal estimation results.
One option is not refer to the PCM in such occlusion environment; however, this approach does not consider the fact that TP results may be presented in the PCM. In the proposed method, we assume that the reliability of the PCM is reduced in occlusion environments.
Thus, we assign a constant weight to the PCM.
The most probable keypoint position is acquired as follows:
\begin{equation}
    _{l}^{t+1}P^n_{key} = \argmax_{-k \leq a,b,c \leq k} \sum_{v}^{n_{v}} \hspace{1mm} {_{l}^{t+1} w_{i}^n} \hspace{1mm} {_l^{t+1}\mathcal S \it _i^n( \it \mu _i ( \hspace{1mm} {_l ^{t+\rm 1 \it }L_{a,b,c}^n))}}
\label{eq:cube_search}
\end{equation}
\begin{equation}
\scalebox{0.85}{$
  _{l}^{t+1} w_{i}^n = \begin{cases}
    g & \mbox{if $\mu _i (_{l}^{t+1} P _{pred}^n)$ is occluded by other $\mu _i (^{t+1} P _{pred}^n)$} \\
    1 & \rm otherwise.
  \end{cases} ,\\
\label{eq:cube_search3}
$}
\end{equation}
\noindent
where $^{t+1}_l \mathcal S \it ^n_i(X)$ represents a function to obtain the PCM value on camera $i$
at time $t+1$ of joint $n$ of person $l$, and $g$ is a constant value in the range [0, 1].

Next, by referencing the probable keypoint position, we compute the joint position of the skeletal model.
With the proposed method, the skeletal model's joint angle is optimized using IK \cite{Ayusawa1} by the keypoint position as the target position while referencing the correspondence shown in Fig. \ref{fig:pcm_figure}.
\begin{equation}
    ^{t+1}_l \bf Q \it = \argmin \sum_n^{n_{k}} \cfrac{\rm 1 \it}{\rm 2 \it} \hspace{1mm} {^{t+\rm 1 \it }_{l} W^n}  ||^{t+\rm 1 \it }_{l} P^n _{key} - ^{t+\rm 1 \it }_{l} P^n ||^{\rm 2 \it}
\label{eq:ik1}
\end{equation}
\begin{equation}
    s.t. \hspace{3mm} {^{t+1}_l \bf \dot{P} \it} = {_{l} \bf J \it } \hspace{2mm} {^{t+1}_{l} \bf \dot{Q}}
\label{eq:ik2}
\end{equation}
\begin{equation}
    ^{t+1}_{l} W^n = \sum_{v}^{n_{v}} {_l^{t+1}\mathcal S \it _i^n( \mu _i ( \hspace{1mm} ^{t+\rm 1 \it }_{l} P^n _{key}))}
\label{eq:ik3}
\end{equation}
\noindent
Here, $^{t+1}_l \bf Q$ represents the joint angle of person $l$ at time $t+1$, ${_{l} \bf J \it }$
represents the Jacobian matrix stands for the forward kinematics, and $^{t+1}_{l} W^n$ is the sum of the PCM value at the probable keypoint position and is used as a weight.

Although joint positions can be computed using the above IK computation, these positions do not consider the temporal continuity of motion.
To obtain smooth motion, the joint position is smoothed using a low-pass filter $\mathcal F$ comprising time-series data of the joint positions.
\begin{equation}
    ^{t+1}_{l} \bf P \it _{smo} = {^{t+\rm 1 \it }_{l} \mathcal F \it}(^{t+\rm 1 \it }_{l} \bf P \it)
\label{eq:lowpass}
\end{equation}

However, when this smoothing procedure is performed, the skeletal structure is collapsed and spatial continuity is lost.
In addition, although only the link length is considered in the above IK computation, each joint angle is expected to not deviate from the RoM. Then, the skeletal model is optimized using IK again by the smoothed joint position as the target position.
\begin{equation}
    ^{t+1}_l \bf Q' \it = \argmin \sum_n^{n_{k}} \cfrac{\rm 1 \it}{\rm 2 \it} \hspace{2mm} ||^{t+\rm 1 \it }_{l} P^n _{smo} - ^{t+\rm 1 \it }_{l} P'^n ||^{\rm 2 \it}
\label{eq:ik2_1}
\end{equation}
\vspace{-0.2cm}
\begin{equation}
\begin{split}
    s.t. \hspace{3mm} & {^{t+1}_l \bf \dot{P}' \it} = {_{l} \bf J \it } \hspace{2mm} {^{t+1}_{l} \bf \dot{Q}'} \\
    & \bf Q \it^{-} \leq {^{t+\rm 1 \it }_l \bf Q'} \it \leq \bf Q \it^{+}
\label{eq:ik2_2}
\end{split}
\end{equation}
\noindent
Here, $\bf Q \it^{-}$ and $\bf Q \it^{+}$ represent the minimum and maximum values of the RoM, respectively \cite{rom:1995}. With the computation above, joint positions and angles with spatiotemporal accuracy
are acquired.

By repeatedly computing the above processes, single-person motion capture is realized, and by computing in parallel to the number of subjects, multi-person video motion capture is realized.

\section{Experimental results}
The proposed method was applied to various datasets as shown in Table \ref{tab:dataset}, including an original dataset, which we refer to as YNL-MP.
For this evaluation, we used three metrics: percentage of correct parts (PCP), percentage of correct keypoints (PCK), and mean per joint position error (MPJPE).
With PCP, a limb is considered detected if the distance between the two calculated joint positions and the true limb joint positions is less than half of the limb length both.
With PCK, a calculated joint is considered correct if the distance between the calculated and true joints is within a certain threshold.
MPJPE represents the average distance between the calculated and true joint positions.
\begin{table}[H]
\centering
\caption{Dataset overview ($n_c$: number of cameras; $n_v$: number of viewpoints; $n_p$: number of persons; $I$: image resolution; $F$: frame rate; $M$: approximate measurement field size [$m^2$]).}
\scalebox{0.9}{
\begin{tabular}{rcccccc}
\hline
Dataset & $n_c$  & $n_v$ & $n_p$ & $I$ & $F$ & $M$\\ \hline
Shelf \cite{shelf}   & 5  & 5 & 2-4 & 1032 $\times$ 776  & 20 & 3 $\times$ 3 \\
YNL-MP\textsuperscript{\ref{note1}}  & 8  & 4 & 1-5 & 1920 $\times$ 1200 & 60 & 5 $\times$ 7 \\
Futsal  & 12 & 4 & 7-8 & 1920 $\times$ 1200 & 60 & 16 $\times$ 24 \\ \hline
\end{tabular}
}
\label{tab:dataset}
\end{table}

In the bone CG in the following figure, the bone length differs for each subject, and the motion is updated according to the calculated joint angle.

\subsection{Evaluation with public dataset}
\begin{figure}[!t]
\centering
\includegraphics[width=0.9\linewidth]{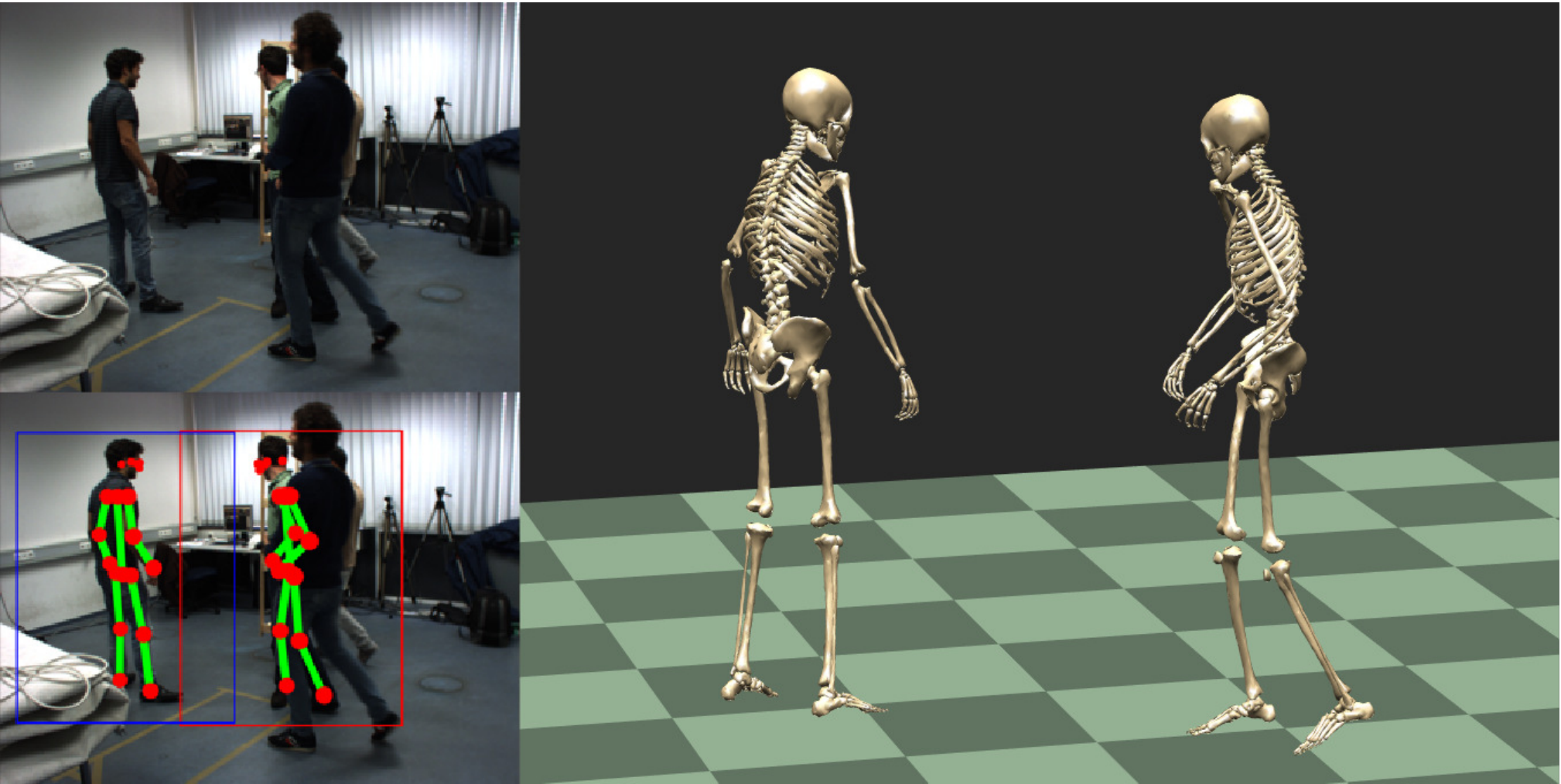}
\caption{Qualitative results obtained on Shelf dataset \cite{shelf}.
\vspace{-0.3cm}
}
\label{fig:shelf}
\end{figure}

The proposed method was applied to the Shelf \cite{shelf} public dataset, in which four people are mutually interacting. These people were recorded using five cameras as shown in Fig. \ref{fig:shelf}.
Here, we employ the same evaluation metrics used in previous studies \cite{Belagiannis2, Ershadi2018, dong2019, Bridgeman}: PCP.

A few points are noteworthy.
First, some subjects are not visible in the initial frame; thus, it is impossible to calculate their initial joint positions and link lengths. Therefore, these subjects were excluded from the analyses.
Second, the ground truth and our skeletal models' joints differ; therefore, only body parts (except for the head) were used to calculate PCP \cite{Bridgeman}.
Third, alternative ways have been used to calculate PCP \cite{dong2019, Bridgeman, 4dAssociation}: a limb is considered detected if the distance between \textit{the midpoint of} two calculated joint positions and \textit{the midpoint of} true limb joint positions is less than half of the limb length. Therefore, we calculate PCP using two ways.
The results are presented in Table \ref{tab:shelf}.
\begin{table}[!t]
\centering
\caption{Comparison of PCP to Shelf dataset \cite{shelf}. The upper part was calculated from
two joint positions constituting limb. The lower part was calculated from the midpoint.
\vspace{-0.3cm}
}
\scalebox{0.95}{
\begin{tabular}{rccc} \hline
Method  & Actor 1 & Actor 2 & Actor 3 \\ \hline
Belagiannis \etal \cite{Belagiannis2}        & 75.3   & 69.7   & 87.6  \\
Ershadi-Nasab \etal \cite{Ershadi2018}       & 93.3   & 75.9   & 94.8  \\
Bridgeman \etal \cite{Bridgeman}             & \textbf{98.8}   & 85.9 & \textbf{97.1} \\
Ours                                         & 98.4   & -      & \textbf{97.1} \\ \hline \hline
Dong \etal \cite{dong2019}                   & 98.8   & 94.1   & 97.8  \\
Bridgeman \etal \cite{Bridgeman}             & 99.7   & 92.8   & 97.7  \\
Zhang \etal \cite{4dAssociation}             & 99.0   & 96.2   & 97.6  \\
Ours                                         & \textbf{99.9} & -   & \textbf{97.9} \\ \hline
\end{tabular}
}
\label{tab:shelf}
\end{table}

The results demonstrate that the proposed method can robustly and accurately reconstruct 3D motion
even in a multi-person environment.
In addition, the proposed method can achieve better or comparable performance than the previous studies \cite{dong2019, Bridgeman, 4dAssociation}.

However, in the dataset, the subject motions are slow and slight. It is too simple to compare with the other state-of-the-art methods, and questionable whether this accuracy can be trusted when used in actual sports scenes.
Therefore, to examine specific problems, e.g., dynamic motion, complex poses, and multiple people, we created an original evaluation dataset to measure multiple subjects.

\subsection{Evaluation with original dataset}
\label{sec:optical}
\begin{figure*}[!t]
\centering
\includegraphics[width=\linewidth]{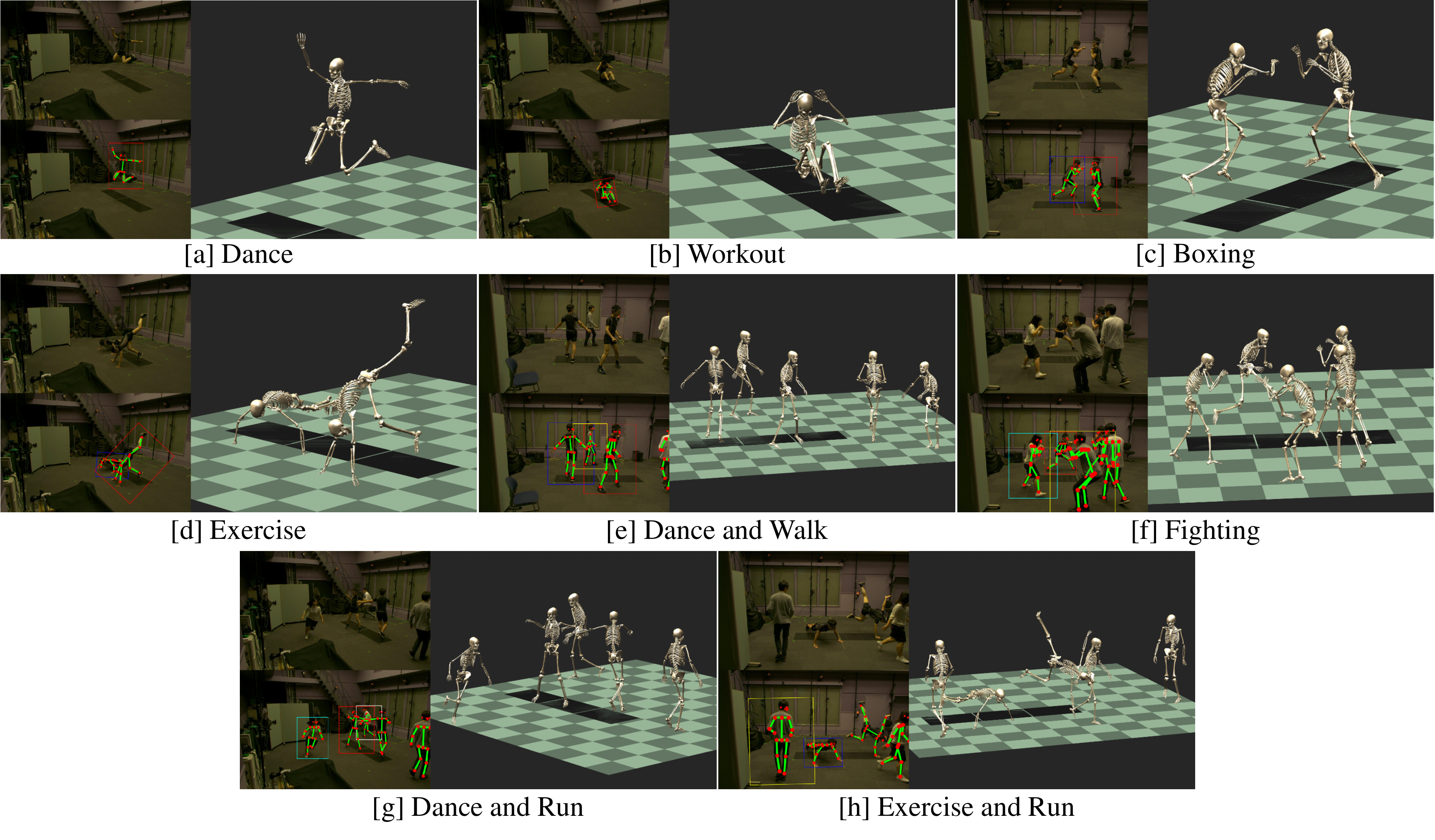}
\caption{Qualitative results obtained on YNL-MP dataset\textsuperscript{\ref{note1}}.
\vspace{-0.3cm}
}
\label{fig:optical}
\end{figure*}

Using eight RGB cameras (acA1920-155uc; Basler AG) at 60 Hz, one to five subjects were recorded.
Also, using 17 infrared cameras (Eagle and Raptor-4; Motion Analysis Corp.) at 200 Hz, two subjects with 44 reflective markers were simultaneously measured.
For this measurement, two RGB cameras were set at each viewpoint to cover the entire measurement field. Eight motions, e.g., boxing, and handstands, were measured as shown in Fig. \ref{fig:optical}.
The dataset will be published with the camera parameters and marker positions for related work\textsuperscript{\ref{note1}}.

\begin{figure*}[!t]
\centering
\includegraphics[width=0.8\linewidth]{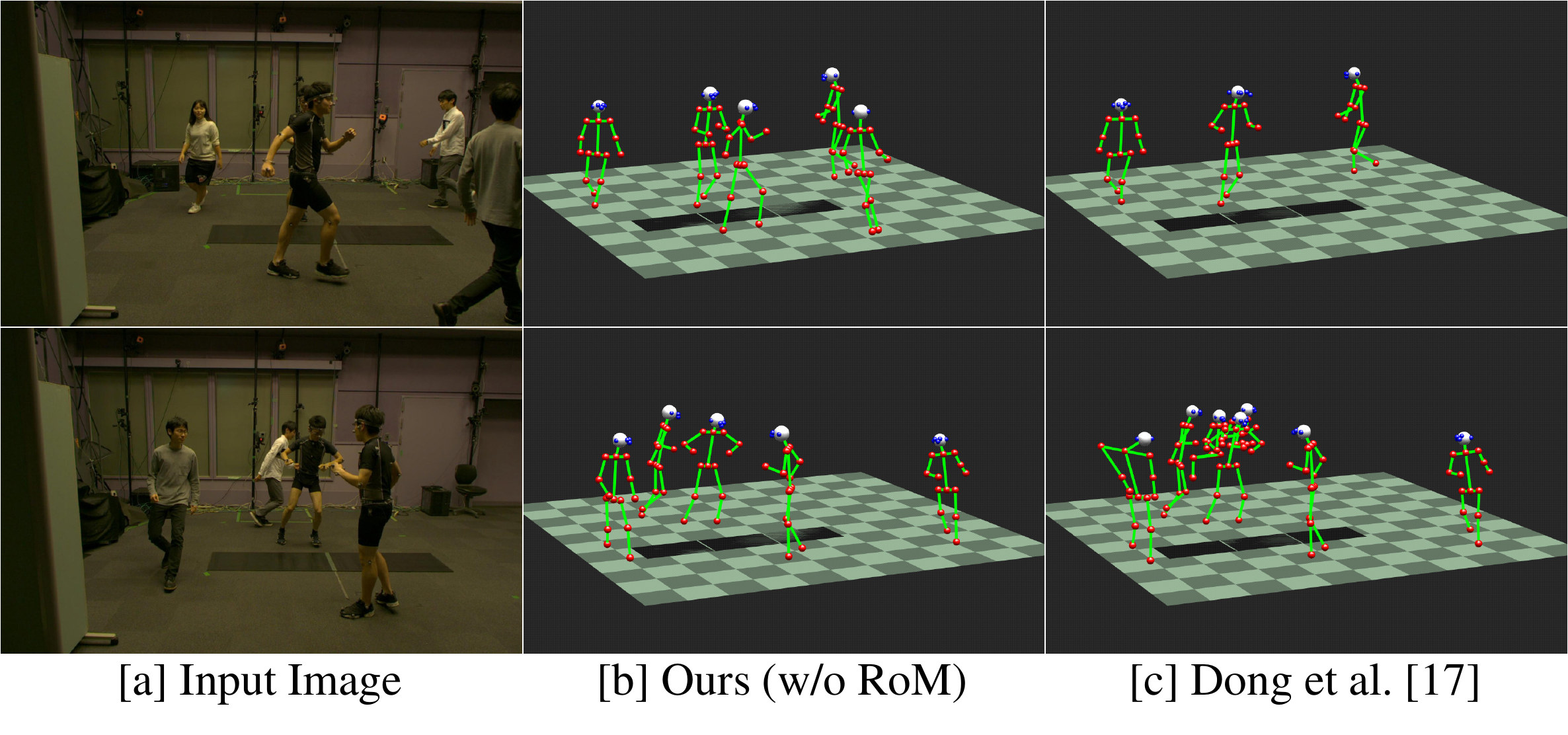}
\caption{Qualitative accuracy comparison between the proposed method and a state-of-the-art method \cite{dong2019} on the Studio dataset\textsuperscript{\ref{note1}}. While the proposed method can always obtain the 3D pose of all the persons in the capture area, the existing method cannot obtain the 3D pose of some persons, or the unintended persons are reconstructed.
\vspace{-0.3cm}
}
\label{fig:mvpose}
\end{figure*}

\begin{figure*}[!t]
\centering
\includegraphics[width=0.8\linewidth]{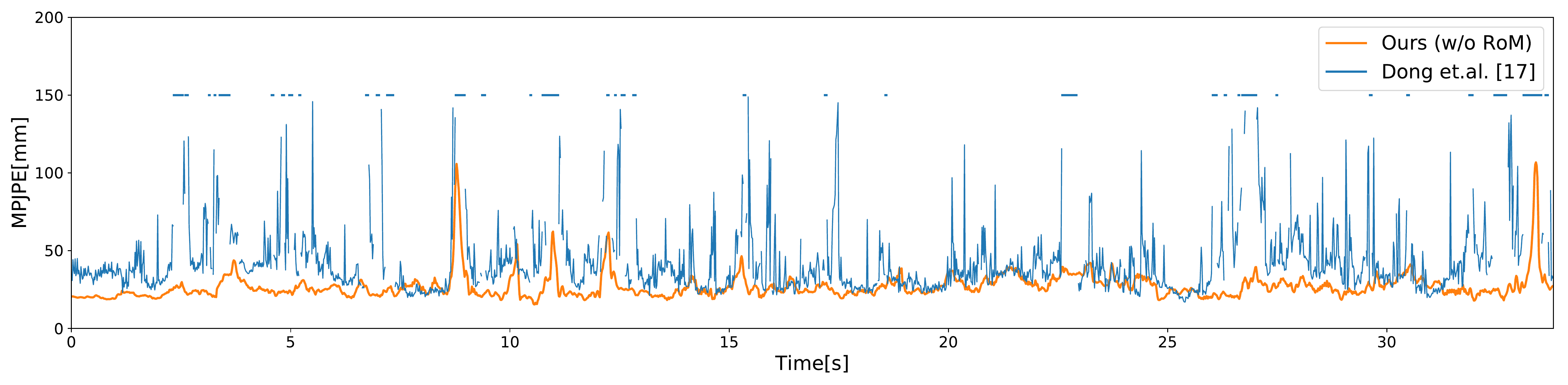}
\includegraphics[width=0.8\linewidth]{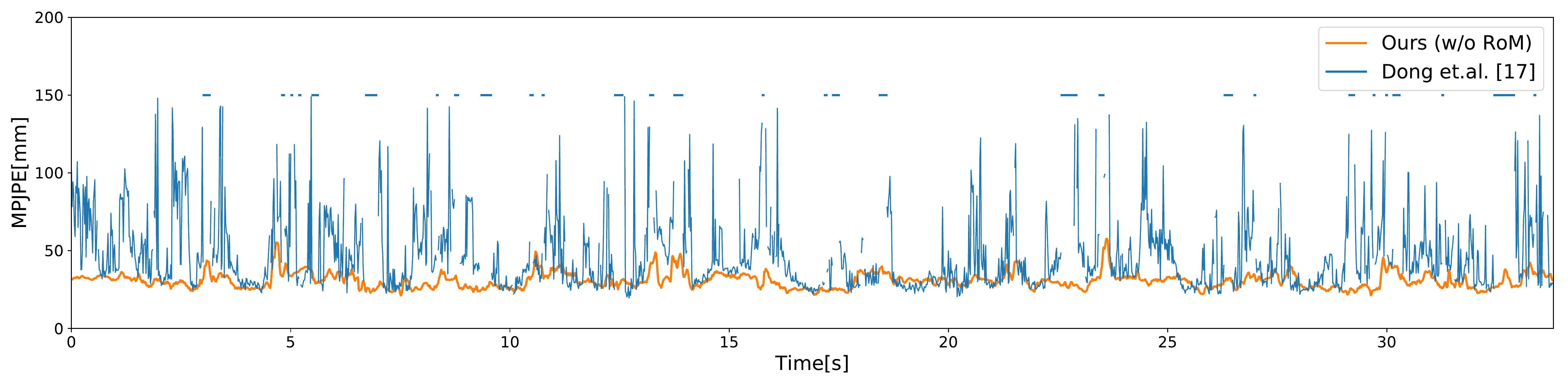}
\caption{
Comparison between the proposed method and a state-of-the-art method \cite{dong2019} for the Studio dataset [g]\textsuperscript{\ref{note1}}. The upper part represents the results of Actor 1, and the lower part represents Actor 2. The proposed method can reconstruct 3D motion in the whole time frame and maintain a low estimation error, whereas the results of the existing method are discontinuous in time and have a large variation in error.
\vspace{-0.3cm}
}
\label{fig:mvpose_graph}
\end{figure*}

Using this dataset, we evaluated the proposed method and a state-of-the-art method with code \cite{dong2019}.
Note that the existing method investigated 3D reconstruction in situations where the number of people in the capture area is unknown.
Thus, depending on the estimation results, the person targeted for reconstruction may be lost or a person who does not exist may be reconstructed by mistake.
Therefore, we defined a new evaluation metric: success rate and.
For a 3D pose whose MPJPE was 150 mm or less compared to the ground truth, it was determined that the 3D pose was successfully reconstructed, and then, the other evaluation metrics were calculated.
The results are summarized in Table \ref{tab:Studio} and shown in Fig.\ref{fig:mvpose} and Fig. \ref{fig:mvpose_graph}.

\begin{table*}[!ht]
\centering
\caption{Evaluation using YNL-MP dataset\textsuperscript{\ref{note1}}.
\vspace{-0.2cm}
}
\scalebox{0.88}{
\begin{tabular}{lccccccccccccccc}
\hline
Dataset           & {[}a{]}  & {[}b{]} & \multicolumn{2}{c}{{[}c{]}}& \multicolumn{2}{c}{{[}d{]}} & \multicolumn{2}{c}{{[}e{]}}& \multicolumn{2}{c}{{[}f{]}}& \multicolumn{2}{c}{{[}g{]}}& \multicolumn{2}{c}{{[}h{]}} & {[}e,f,g{]} \\ \hline
Number of Persons  & 1    & 1   & \multicolumn{2}{c}{2}  & \multicolumn{2}{c}{2}   & \multicolumn{2}{c}{5}  & \multicolumn{2}{c}{5}  & \multicolumn{2}{c}{5}  & \multicolumn{2}{c}{5}   &    5    \\
Total Time{[}s{]} & 35.2 & 32.8&\multicolumn{2}{c}{28.2}& \multicolumn{2}{c}{30.8}&\multicolumn{2}{c}{33.8}&\multicolumn{2}{c}{30.9}&\multicolumn{2}{c}{34.2}& \multicolumn{2}{c}{30.1}&   98.9  \\
Need Rotation?    & No   & Yes & \multicolumn{2}{c}{No} & \multicolumn{2}{c}{Yes} & \multicolumn{2}{c}{No} & \multicolumn{2}{c}{No} & \multicolumn{2}{c}{No} & \multicolumn{2}{c}{Yes} &    No   \\
Actor ID          & 1    & 1   & 1          & 2         & 1          & 2          & 1          & 2         & 1          & 2         & 1          & 2         & 1          & 2        & Ave \\ \hline \hline
Ours                   &      &      &             &            &             &            &             &            &             &            &             &            &             &            &             \\
Success Rate@150mm & 100 & 100 & 100        & 100       & 100        & 100       & 100        & 100       & 100        & 100       & 99.9        & 100       & 100        & 100       & 100        \\
MPJPE {[}mm{]}           & 27.5 & 38.6 & 28.8        & 32.9       & 36.3        & 40.0       & 29.2        & 32.6       & 31.1        & 33.1       & 30.8        & 32.2       & 32.8        & 51.2       & 31.5        \\
PCP                      & 100  & 98.2 & 99.4        & 99.6       & 97.7        & 99.0       & 99.7        & 99.8       & 99.4        & 99.6       & 98.6        & 99.8       & 99.0        & 98.6       & 99.5        \\
PCK@50mm                 & 96.3 & 75.8 & 93.8        & 88.4       & 80.9        & 70.9       & 93.3        & 85.3       & 89.2        & 88.1       & 91.3        & 87.7       & 86.3        & 56.4       & 89.2        \\
PCK@100mm                & 99.9 & 99.4 & 99.5        & 99.6       & 99.0        & 98.9       & 99.5        & 99.8       & 99.3        & 99.5       & 98.6        & 99.8       & 99.3        & 95.2       & 99.4        \\ \hline
Ours (w/o RoM)           &      &      &             &            &             &            &             &            &             &            &             &            &             &            &             \\
Success Rate@150mm & 100 & 100 & 100        & 100       & 100        & 100       & 100        & 100       & 100        & 100       & 100        & 100       & 100        & 100       & 100        \\
MPJPE {[}mm{]}           & 25.4 & 36.9 & 27.3        & 30.3       & 35.3        & 37.5       & 27.0        & 30.8       & 29.5        & 30.7       & 27.8        & 30.3       & 31.8        & 50.3       & 29.3        \\
PCP                      & 100  & 98.4 & 99.5        & 99.5       & 97.9        & 99.2       & 99.7        & 99.8       & 99.4        & 99.5       & 98.6        & 99.8       & 99.0        & 98.8       & 99.5        \\
PCK@50mm                 & 98.0 & 79.9 & 95.0        & 91.0       & 82.4        & 76.6       & 95.8        & 90.1       & 91.0        & 92.0       & 94.1        & 92.4       & 87.2        & 59.2       & 92.6        \\
PCK@100mm                & 99.9 & 99.5 & 99.6        & 99.6       & 98.9        & 99.3       & 99.5        & 99.8       & 99.2        & 99.5       & 98.6        & 99.8       & 99.2        & 95.4       & 99.4        \\ \hline
Dong \etal \cite{dong2019} &    &      &             &            &             &            &             &            &             &            &             &            &             &            &             \\
Success Rate@150mm & 90.8 & 87.1 & 97.4        & 94.6       & 88.6        & 97.3       & 86.1        & 87.1       & 86.6        & 84.5       & 87.9        & 81.1       & 87.5        & 69.0       & 85.5        \\
MPJPE {[}mm{]}     & 62.3 & 56.1 & 36.7        & 43.7       & 41.3        & 58.5       & 41.7        & 46.8       & 51.4        & 48.2       & 43.2        & 49.1       & 48.4        & 88.2       & 46.6        \\
PCP                & 78.0 & 80.1 & 94.0        & 90.1       & 84.3        & 87.6       & 82.1        & 80.4       & 78.9        & 79.3       & 82.4        & 74.1       & 81.5        & 55.4       & 79.5        \\
PCK@50mm           & 57.1 & 50.2 & 82.7        & 74.6       & 71.4        & 57.5       & 69.9        & 68.5       & 62.5        & 62.3       & 69.8        & 61.9       & 67.7        & 29.3       & 65.9        \\
PCK@100mm          & 75.8 & 78.7 & 94.9        & 89.6       & 83.8        & 86.6       & 81.4        & 80.3       & 78.6        & 79.2       & 82.3        & 74.4       & 80.9        & 51.9       & 79.4        \\ \hline
\end{tabular}
}
\label{tab:Studio}
\end{table*}
Results demonstrate that the proposed method can reconstruct 3D motion robustly and accurately, even in five-person environments as well as in single-person environments.
The proposed method achieved 31.5 mm in MPJPE and 99.5\% in PCP for five-person dynamic movement. These results indicate that the proposed method achieved better or comparable performance for a single-person environment than a previous study \cite{Ohashi:2018} (26.1 mm in MPJPE and 95.8\% in PCK@50 mm without RoM).
In addition, even in a challenging environment in which the human pose was significantly inclined, e.g., handstands or push-ups, which are generally difficult for pose estimation, the 3D motion can be acquired by rotating the bounding box. In such cases, the proposed method achieved greater than 95.0\% in PCP in a five-person environment.

The bone CG in Fig. \ref{fig:optical} shows that the proposed RoM restriction works under
dynamic motion, thereby preventing strange pose reconstruction. In addition, the \textit{supplemental video}\textsuperscript{\ref{note1}} shows that the proposed method can
draw CG without causing feet-sliding: generally caused by fitting the 3D keypoint position with CG model which has different scale size.
However, this restriction can have an adverse effect: when performing dynamic motion, e.g., swinging the arms, optimization may fall into a singular posture; thus, optimal joint positions cannot be acquired.
In Table \ref{tab:Studio}, comparing the results obtained with and without RoM reveals that the latter achieves higher accuracy. Therefore, if only 3D joint positions are required, the RoM is not expected to be restricted. However, if, for example, the motion data are used for CG production or medical diagnosis, then 3D reconstruction with RoM is more suitable.

The comparison of the proposed method to the existing method \cite{dong2019} demonstrates that the proposed method is superior in terms of accuracy, and only the proposed method can estimate a temporally continuous 3D motion.
Thus, we consider that the proposed method is more suitable than the 2D keypoint triangulation method, the existing method's framework, for multi-person markerless motion capture.

\subsection{Experiment on futsal field}
\begin{figure*}[!t]
\begin{center}
\includegraphics[width=\linewidth]{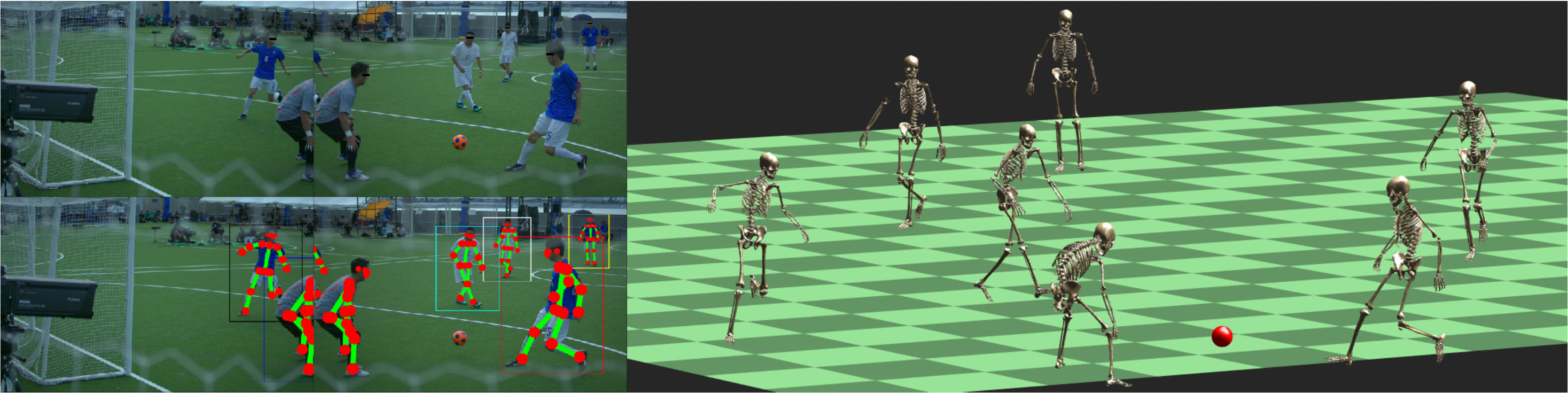}
\caption{Qualitative result on futsal field.
\vspace{-0.3cm}
}
\label{fig:out_put}
\end{center}
\end{figure*}

We measured futsal games to verify the proposed method in a real-world environment.
In the measurement, to cover approximately two-thirds of the court with the camera's field of view, 12 RGB cameras were set at four corners, and eight players were recorded.
The futsal ball was detected by color from each camera, and reconstructed in 3D.
As an aside, using the ball trajectories, bundle adjustment \cite{Bill:2000} was performed, and camera parameters were acquired.
The results are presented in Fig. \ref{fig:out_put}.

Note that no ground truth was available; thus, the results represent a qualitative evaluation.
However, the results of re-projected joint positions onto the input image and the bone CG demonstrate that motion capture can be achieved with accuracy nearly equal to that of experiments in \textit{Section} \ref{sec:optical}.
Using only a few cameras, all players' detailed motion was successfully acquired.

\section{Conclusion}
The conclusions obtained from this study are following.
\vspace{-0.1cm}
\begin{enumerate}
    \item A method to realize multi-person motion capture using multiple video cameras was proposed by predicting accurate 3D pose and a bounding box. The proposed method works even in a wide field using cameras with different fields of view placed at a single viewpoint.
    \item By considering link length, RoM, and spatiotemporal continuity of motion, accurate and smooth motion data can be obtained.
    \item The proposed method achieved 31.5 mm in MPJPE and 99.5\% in PCP in an environment with five people dynamically moving while satisfying the RoM.
    \item With the proposed method, all players' detailed motions in a futsal game were acquired only from a few cameras.
\end{enumerate}
\vspace{-0.1cm}

Our approach still has limitations.
In the proposed method, individual pose estimation is performed with the bounding box. However, when two subjects are extremely close, e.g., when hugging, the pose estimator cannot compute the PCM of the intended subject from every camera, which leads to failure.
Furthermore, when the subject completely moves out of the sight of the two cameras, e.g., when the subject is completely occluded or comes too close to the camera, reconstruction cannot be performed.
However, we hope this work will guide future realization of multi-person markerless motion capture in more challenging environments, e.g., real soccer matches.

\section*{Acknowledgements}
This work was made using sDIMS, a programming library for multi-body kinematics and dynamics with the human musculo-skeletal model developed in the University of Tokyo. The authors acknowledge the supports by Ayaka Yamada, Hiroki Obara, Tomoyuki Horikawa and the other students in the futsal motion capture experiment. We also thank the anonymous participants in the studio motion capture experiment. This work was conducted in the research funded by JSPS Grants-in-Aid for Scientific Research (A) JP17H00766 (2017-2019) and by NTT DOCOMO, Inc.

\renewcommand{\theequation}{A.\arabic{equation}}
\setcounter{equation}{0}

\appendix
\section*{Appendix}
\renewcommand{\thesubsection}{\Alph{subsection}}
\subsection{Camera calibration in wide field}

A $3 \times 4$ matrix $M_{i}$ to project an arbitrary 3D point onto the image plane of camera $i$ is expressed as follows:
\begin{equation}
    M_{i} \equiv K_{i} \begin{bmatrix} R_{i} | \bf t_{\it i} \end{bmatrix}
\label{eq:cam_matrix}
\end{equation}
\noindent
where $K_{i}$ is an internal parameter, and $R_{i}$ and $\bf t_{\it i}$ are external parameters representing the attitude and position of the camera, respectively.
Here, the distortion parameter can be calculated together with the internal parameter; thus, in the following, it is assumed that the internal and distortion parameters are calculated using the
chess pattern \cite{Zhang:2000}, and the input image is compensated in advance.

The external parameters are acquired using the Structure from Motion (SfM) approach \cite{Hartley:2003, Joel:2003} as follows.
\begin{enumerate}
    \vspace{-0.2cm}
    \item The cameras are set at each viewpoint, and the external parameters of each camera are roughly estimated.
    \vspace{-0.2cm}
    \item A colored sphere is moved to cover the measurement area. Then, the center of the sphere is detected from multiple synchronized cameras. Reconstruct them in 3D by triangulation while removing the outlier using RANSAC.
    \vspace{-0.2cm}
    \item Using bundle adjustment, the attitude and position of the cameras and 3D positions of the sphere are optimized \cite{Bill:2000}. With this method, we treat the rotation matrix, translation vector, and focal length as variables and then apply the Ceres Solver for bundle adjustment \cite{ceres-solver}.
    \vspace{-0.2cm}
    \item The absolute position, attitude, and scale to world coordinates are transformed while maintaining the relative relation between cameras.
    \vspace{-0.2cm}
\end{enumerate}

Camera calibration is performed using the process described above. A projection matrix $M_{i}$ is obtained from each camera. The pixel position where point $X$ is projected onto the image plane of camera $i$ is expressed as follows.
\begin{equation}
    \mu _i (X) =
    \begin{pmatrix}
        \begin{bmatrix} M_{i} X \end{bmatrix}_x / \begin{bmatrix} M_{i} X \end{bmatrix}_z \\
        \begin{bmatrix} M_{i} X \end{bmatrix}_y / \begin{bmatrix} M_{i} X \end{bmatrix}_z
    \end{pmatrix}
\label{eq:projection}
\end{equation}

\subsection{Skeletal model and joint position initialization}
\label{sec:Initialization}
To compute IK \cite{Ayusawa1}, the skeletal model's adjacent joints must be connected by a constant-length link.
Here, link length must be calculated according to the human subject.
In addition, IK is based on iterative computation; therefore, it is reasonable to calculate the skeletal model's initial joint position before IK computation.
In the proposed method, using multi-camera images, the pixel locations of the keypoint detected from HRNet \cite{Xiao:2018, Sun:2019} at an initial frame are reconstructed in 3D.
The length parameters and initial joint position are simultaneously calculated from the 3D keypoint positions.

The initial bounding box position is roughly calculated by the 3D keypoint positions reconstructed by the 2D keypoint positions detected by the bottom-up pose estimator \cite{higherHRNet} while considering the epipolar constraints, or given manually.
In addition, the number of keypoints is less than the number of skeletal model's joints; thus, at the initial frame, restrictions, e.g., unbent spine and not-raised scapula, are added to the subjects. Further, the parameters are restricted such that the left and right lengths are symmetrical.

{\small
\bibliographystyle{ieee_fullname}
\bibliography{myreference}
}
\end{document}